\documentclass[runningheads]{llncs}
\usepackage{graphicx}

\usepackage{amsmath}
\usepackage{amssymb}
\usepackage{cite}

\DeclareMathOperator{\EX}{\mathbb{E}}

\begin{document}
\title{Modular Networks Prevent Catastrophic Interference in Model-Based Multi-Task Reinforcement Learning}
\titlerunning{Modular Networks Prevent Catastrophic Interference in MB Multi-Task RL}
%
\author{Robin Schiewer\inst{1}\orcidID{0000-0002-4021-2476}\and
Laurenz Wiskott\inst{1}\orcidID{0000-0001-6237-740X}}
\authorrunning{R. Schiewer \and L. Wiskott}
%
\institute{Institute for Neural Computation, Ruhr-University Bochum, Bochum, Germany\\
\email{\{firstname.lastname\}@ini.rub.de}\\
\url{https://www.ini.rub.de/}
}
\maketitle              
\begin{abstract}
In a multi-task reinforcement learning setting, the learner commonly benefits from training on multiple related tasks by exploiting similarities among them. At the same time, the trained agent is able to solve a wider range of different problems. While this effect is well documented for model-free multi-task methods, we demonstrate a detrimental effect when using a single learned dynamics model for multiple tasks. Thus, we address the fundamental question of whether model-based multi-task reinforcement learning benefits from shared dynamics models in a similar way model-free methods do from shared policy networks. Using a single dynamics model, we see clear evidence of task confusion and reduced performance. As a remedy, enforcing an internal structure for the learned dynamics model by training isolated sub-networks for each task notably improves performance while using the same amount of parameters. We illustrate our findings by comparing both methods on a simple gridworld and a more complex vizdoom multi-task experiment.

\keywords{model-based reinforcement learning \and multi-task reinforcement learning \and latent space models \and catastrophic interference \and task confusion.}
\end{abstract}
\section{Introduction}

In recent years, deep reinforcement learning (RL) has shown impressive results in problem domains such as robotics and game playing \cite{Mnih2013, Mnih2016, Hessel2017, Schulman2017, Haarnoja2018}. However, sample inefficiency is still a major shortcoming of many of the methods. To achieve superhuman performance e.g. in video games, the required number of interactions for complex tasks lies in the tenths of millions. Model-based approaches mitigate this problem by integrating the collected sample information into a coherent model of the environmental dynamics \cite{Sutton1991}. Those learned models are used for direct policy learning \cite{Ha2018}, planning \cite{Hafner2018, Ebert2018} or to augment existing model-free approaches \cite{Feinberg2018, Weber2017, Nagabandi2017}. In the presence of a readily available analytical model of the environment, that does not have to be learned, perfect planning can be used to learn a policy with great success \cite{Tesauro1995, Silver2018}. 

Despite the success of deep RL across a wide range of problem domains, most state of the art approaches have to be re-trained for every new task. Since similar problems might very well share a common underlying structure, this is a potential waste of resources \cite{Caruana1997}. Furthermore, an agent can also encounter a different mixture of tasks in a single environment. If the agent is created without the concept of distinct tasks in mind, an alteration of the task mixture may negatively impact the agent's performance. While greater sample efficiency is one reason to engage in multi-task learning, it also produces a more capable agent that can solve a wider range of problems. Whereas there exists a vast amount of research around multi-task RL in general \cite{Rusu2016, Fernando2017, Finn2017, Duan2016}, the combination of model-based latent space RL and multiple tasks is still largely unexplored. Since the former has brought impressive improvements for sample efficiency and the latter promises greater flexibility and reusability, we combine latent space models with multi-task RL in this paper and investigate its usefulness.

We focus on model-based multi-task deep RL for complex observation domains like images. The main question we address is if training the same dynamics model on multiple similar tasks helps performance through knowledge transfer or if catastrophic interference outweighs the benefits. If so, how can a model-based multi-task agent be structured to avoid catastrophic interference? Our contributions in answering these questions are the following:

\begin{itemize}
    \item In our experiments we show a detrimental effect on performance through catastrophic interference when training a single dynamics model on multiple tasks simultaneously. To mitigate the effect, we propose a world model that uses multiple distinct latent space dynamics models which are activated through a context-based task classification network.
    \item By strict separation of the dynamics networks, the probability that dynamics of different tasks interfere with each other is minimized. As a side effect, retraining individual tasks and recombining learned tasks is straightforward.
    \item We demonstrate the performance difference between a single dynamics model and our method in a 3-task gridworld and a more complex 2-task 3D environment. To evaluate both approaches, we perform planning on the learned dynamics models and measure the obtained reward.
\end{itemize}

\section{Related Work}

\subsubsection{Latent Space Models:} Previous work on learning latent space models from high dimensional inputs has produced impressive results especially in the atari learning environment and continuous control benchmarks \cite{Oh2015, Hafner2020, Kaiser2019}. The architectures share a similar structure where a convolutional encoder embeds inputs to a latent representation which is fed into a recurrent prediction network for states and rewards. By stacking predictions on previous predictions, imaginary rollout trajectories based on the learned dynamics are produced. Those are used for planning \cite{Hafner2018}, training an actor-critic agent \cite{Hafner2018, Hafner2020} or to drive exploration \cite{Sekar2020}. To mitigate accumulating errors caused by imperfect dynamics models, probabilistic prediction models and long rollout trajectories during training are used. While the above mentioned methods demonstrate impressive performance of latent space models in RL, we specifically focus on resource efficiency in a multi-task setting.

\subsubsection{Multi-Task RL:} To overcome catastrophic interference and pave the way for lifelong learning, in \cite{Rusu2016} a policy network from multiple modules called columns is built. Each column is a mini network with individual weights while the same architecture is shared among all columns. For each task, a new column is added and layer-wise laterally connected to all previous columns. Since the connections are learned, this enables new columns to selectively benefit from previously acquired behavior. The authors train their method using A3C \cite{Mnih2016} and demonstrate performance improvements by comparing it to various versions of a single-column baseline network. 

By learning task-specific pathways of active subsections within one large neural network, \cite{Fernando2017} use a similar concept than \cite{Rusu2016}. A subsection corresponds to a small, localized region in the larger network. However, instead of hardwiring the connections between subsections, they are learned by an evolutionary strategy. Given an active path, training of the involved subsections is done via gradient descent. At the same time, the remaining weights are frozen so that change only occurs in the currently active modules. The authors describe their approach as an evolutionary version of dropout, where thinned out networks are evolved instead of randomly generated (as in regular dropout).

A significant difference between the above mentioned approaches and our method is that our algorithm learns modular dynamics models instead of modular policies. Furthermore, in our proposed method lateral connections between the models are absent since they would introduce dependencies which prevent arbitrary recombination of the learned task models.

\section{Method Description}

\begin{figure}
\centering
\includegraphics[width=\textwidth]{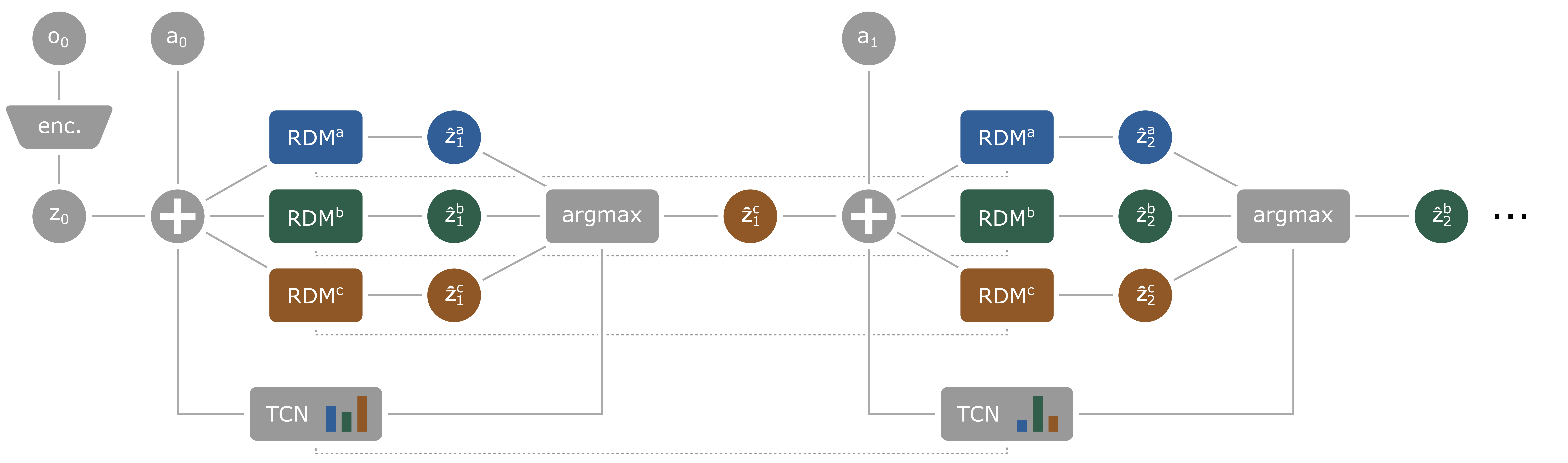}
\caption{Overview of the multi RDM architecture. In the first timestep, observation $o_0$ is embedded by the encoder of the VQ-VAE. The resulting embedding is combined with the action and presented to the RDMs and the TCN. The latter chooses one of the RDMs (visualized by the argmax node), which is consequently used to predict the embedding for the next timestep. For simplicity, reward and terminal state probability outputs are not visualized here. Dashed lines indicate recurrent components that maintain information persistent between the individual time steps.}
\label{fig:model-overview}
\end{figure}

The foundation of our approach lies in the combination of multiple recurrent dynamics models (RDMs), a vector-quantized variational autoencoder (VQ-VAE) and a task classification network (TCN) for online task detection. The VQ-VAE is used to encode the image observation stream from the environment to a latent space representation which we simply refer to as embedding from now on. The TCN receives the stream of embeddings and predicts the probability of each RDM to be responsible for the task at hand. The most probable RDM is then used to generate predictions in latent space. A graphical overview of the method is presented in Figure \ref{fig:model-overview}. In the following sections, we will explain each of the components in more detail. While we discuss the hyperparameters we use for some parts of the architecture, we do not mention all of them and refer to the link to our code repository in the conclusion for an exhaustive list.

\subsection{Vector-Quantized Variational Autoencoder}

To embed an image observation into a latent space, we use the Vector-Quantized Variational Autoencoder (VQ-VAE) from \cite{Oord2017}. It consists of a convolutional encoder block, a quantization layer and a deconvolutional decoder block. In the following, we summarize the parameters of all VQ-VAE components with the parameter vector $\phi$. The encoder transforms an image observation $\mathbf{o}$ into a 3D tensor $E_\phi(\mathbf{o})$ of size $w \times h \times c$. It can be interpreted as a more compact image of size $w$ times $h$ with $c$ channels. This image tensor is passed to the quantization layer that replaces each pixel along the channel dimension with the closest from a set of codebook vectors. The resulting tensor $\mathbf{z}$ is a per-pixel quantized version of $E_\phi(\mathbf{o})$. Conceptually equivalent to a regular VAE, the deconvolutional decoder $D$ transforms $\mathbf{z}$ back to the (reconstructed) original input image $\mathbf{\hat{o}} = D_\phi(\mathbf{z})$. The VQ-VAE is trained by minimizing the following loss function:

\begin{equation}
    \mathcal{L}_\phi(\mathbf{o}, D_\phi(\mathbf{z})) = \underbrace{|| \mathbf{o} - D_\phi(\mathbf{z})||^2_2}_{\text{decoder}} + \underbrace{||sg[E_\phi(\mathbf{o})] - \mathbf{z}||^2_2}_{\text{codebook}} + \beta \underbrace{||sg[\mathbf{z}] - E_\phi(\mathbf{o})||^2_2}_{\text{encoder}}
    \label{eqn:vq-vae-loss}
\end{equation}

Whereas the parameter $\beta$ trades off encoder vs.\ decoder loss and the $sg[\cdot]$ operator indicates the stop of gradient backpropagation. We used $\beta = 0.5$ in all our experiments, since it performed best among a collection of tested values between $0.25$ and $0.75$. According to the recommendations of \cite{Oord2017}, we swap the codebook loss in equation \ref{eqn:vq-vae-loss} for an exponentially moving average update of the codebook vectors $e_i$:

\begin{equation*}
    N_i^{(t)} := N_i^{(t-1)} \gamma + n_i^{(t)} (1 - \gamma), \hspace{1em} m_i^{(t)} := m_i^{(t-1)} \gamma + \sum_j^{n_i^{(t)}} E(o)_{i,j}^{(t)} (1 - \gamma), \hspace{1em} e_i^{(t)}:= \frac{m_i^{(t)}}{N_i^{(t)}}
\end{equation*}
\noindent
where $n_i^{(t)}$ is for the current batch the number of pixels/vectors in $E(\mathbf{o})$ that will be replaced with codebook vector $e_i$ and $\gamma$ is a decay parameter between 0 and 1. We used the default $\gamma=0.99$ for all experiments and a Keras \cite{tensorflow, keras} port of the VQ-VAE implementation in the Sonnet library \cite{sonnet}. The autoencoder architecture of our choice naturally forms clusters in embedding space, which has been reported to increase prediction performance of dynamics models like ours \cite{Oord2017} and potentially helps the TCN in differentiating between tasks.

\subsection{Recurrent Dynamics Models}

For the design of our RDMs, we follow ideas from \cite{Hafner2018} and combine probabilistic and deterministic model components to increase predictive performance and prevent overfitting \cite{Chua2018}. As previously explained, following the work in \cite{Oh2015} we additionally embed environment observations into latent space representations $z$ and operate entirely in that latent space from there on. Since $z$ is a 3D tensor, the architecture of our RDM contains mainly convolutional layers (conv) to carry through any spatial information contained in the image observations. An RDM is comprised of four submodules: The belief-state network (conv), the embedding network (conv), the reward network (dense) and the terminal state probability network (dense).

As already briefly mentioned, to make the prediction of $z_t$ more robust to uncertainty, we combine a stochastic and a deterministic prediction path. The deterministic path is realized through a convolutional LSTM network which computes the belief state $h_t$ from the previous belief state $h_{t-1}$, action $a_{t-1}$ and embedding $z_{t-1}$. Given $h_t$, the stochastic prediction path is a categorical distribution from which $z_t$ is sampled. This way, information about past observations, actions and belief states can be accumulated in $h_t$ and passed on to future time steps. At the same time, uncertainty regarding the next $z_t$ can be expressed through the sampling process. Because the embeddings are defined completely by the indices of their codebook vectors, we use a categorical distribution to sample those indices, namely the Gumbel Softmax \cite{jang2017, maddison2017}, which is a reparameterized approximation to a categorical one-hot distribution. It can be used with backpropagation and at the same time offers almost discrete sampling of the required one-hot vectors. The rewards are sampled from a diagonal Gaussian and the terminal transition probabilities are sampled from a Bernoulli distribution. The structure of the RDM is summarized as follows:

\begin{align}
    \text{belief state:} \ & h_t = f_\theta(h_{t-1}, z_{t-1}, a_{t-1}) \\
    \text{embedding:} \ & z_t \sim q_\theta(z_t \vert h_t) \\
    \text{reward:} \ & r_t \sim q_\theta(r_t \vert h_t) \\
    \text{terminal transition:} \ & \gamma_t \sim q_\theta(\gamma_t \vert h_t)
\end{align}

\noindent
Note that we summarize all RDM component's parameters into one vector $\theta$ for simplicity. To train the RDM, the following loss function is minimized per time step:

\begin{equation}
    \mathcal{L}_\theta = \mathcal{L}_\theta^{e} + \mathcal{L}_\theta^{r} + \mathcal{L}_\theta^{\gamma}
    \label{eqn:rdm-loss-abstract}
\end{equation}

\noindent
whereas the individual prediction loss terms are the parameter maximum likelihood solutions given the transition data:

\begin{align}
    \text{embedding}: \ & \mathcal{L}_\theta^{e} = \EX_{q_\phi(z_t \vert o_t)} \left[ - \ln q_\theta(z_t \vert h_t) \right] \\
    \text{reward}: \ & \mathcal{L}_\theta^{r} = \EX_{q_\theta(z_t \vert h_t), p(r_t \vert o_{t}, a_t, o_{t+1})} \left[ - \ln q_\theta(r_t \vert h_t, z_t) \right] \\
    \text{terminal}: \ & \mathcal{L}_\theta^{\gamma} = \EX_{q_\theta(z_t \vert h_t), p(\gamma_t \vert o_{t}, a_t, o_{t+1})} \left[ - \ln q_\theta(\gamma_t \vert h_t, z_t) \right]
\end{align}

\noindent
while we denote all learned distributions by $q_\theta(\cdot)$ or $q_\phi(\cdot)$ and the true environment dynamics by $p(\cdot)$.

\subsection{Context Detection}

In contrast to the algorithms presented in \cite{Hafner2018, Ha2018, Oh2015}, we use distinct dynamics models to combat catastrophic interference when training on multiple tasks. In theory, a single (monolithic) dynamics model can exploit similarities in the tasks it learns, which positively influences learning speed and predictive performance. But since weight updates for different tasks may interfere with each other negatively, a monolithic dynamics model could instead suffer from reduced performance. This effect has been shown by \cite{Yu2019} for policy networks in multi-task reinforcement learning. To prevent this catastrophic interference, we isolate knowledge about different task dynamics in separate RDMs and use the TCN to orchestrate them. This essentially makes the TCN a context detector that learns to classify an embedding-action-stream and choose the correct RDM for the task at hand. In every prediction step, the TCN receives the current observation embedding and action as input. While it is theoretically able to choose a different RDM in every time step, we did not observe this behavior in practice. Because different tasks can have locally similar or equivalent embeddings, the TCN contains convolutional LSTM layers and is thus able to remember past embeddings and actions. It is thereby capable to disambiguate temporarily similar embedding streams. The TCN is trained in a supervised manner via categorical crossentropy loss.

\subsection{Planning \label{sec:planning}}

We directly use our learned RDMs for planning in the environments. Thereby, we follow the crossentropy method that has originally been introduced by \cite{Rubinstein1997} and is also employed by \cite{Ha2018} and \cite{Chua2018}. A planning procedure consists of $n$ iterations. At the beginning of the first iteration, the current observation is embedded into latent space by the VQ-VAE and serves as a starting point for $k$ rollouts of length $T$. To generate the action sequences required for the rollouts, we initialize one categorical distribution per time step at random from which we sample $k$ actions. This results in a $k \times T$ matrix of actions where each row represents the action sequence for one rollout trajectory. The resulting trajectories produced by the monolithic or multi RDM are ranked w.r.t. their discounted return and the top $\rho$ percent are chosen as winners. In the next iteration, we use the maximum likelihood parameters of the previous winners for the action distributions and again sample $k \times T$ actions to generate new rollouts. For the gridworld setting, we use $n = 20$ iterations, $k = 500$ rollout trajectories and $T = 100$ planning steps, because especially in the third task the reward is far away from the starting location. We take the top $\rho = 10 / 500$ of all rollout trajectories in each iteration. For the 3D environment, we use $n = 5$ iterations, $k = 400$ rollout trajectories and $T = 60$ time steps due to technical limitations regarding the capacity and of the used GPUs (the algorithm is running on a single Tesla v100 with 16 GB of VRAM). We take the top $\rho = 10 / 400$ of all rollout trajectories in each iteration. For both the gridworld and the 3D environment, we add noise of $0.05$ to the maximum likelihood parameters during planning because this effectively improved the results.

\subsection{Training \label{sec:training}}

Presented with a batch of encoded trajectories from $m$ tasks, the TCN outputs the probabilities for each of the $m$ RDMs per sample (i.e. per trajectory per time step). Since the individual prediction errors of the RDMs are weighted with their probability of being chosen, only those with a high probability receive a strong learning signal and consequently adapt their weights. To discourage the TCN from switching around RDMs during a task, we provide the task ID as a supervised learning signal during training. The TCN produces a less concentrated probability distribution in the beginning of the training when the prediction error of the RDMs dominates the loss. At this time, multiple RDMs can be chosen for one task without hurting the loss significantly, which means that most RDMs are exposed to more than one task during the course of the training. This increases variety in training samples for each of the RDMs and encourages them to learn characteristics shared across tasks. However, this situation only lasts for the first few training epochs. As soon as the prediction error decreases, the TCN has to allocate the correct models to their respective tasks in order to further minimize the loss. The resulting system is then able to activate a task-specific RDM for each task it was trained on. As a consequence of using distinct networks for each task, individual RDMs may be re-trained or trained longer on tasks where the performance is not yet as desired without the risk of sacrificing performance in other tasks. 

We compare the multi RDM model with a larger, monolithic RDM in two different experiments: A simple 3-task gridworld and a more demanding 2-task setting comprised of two vizdoom \cite{wydmuch2018} environments. The latter setting can be considered more demanding since it features a dynamic 3D environment, more diverse observations and a more complex reward function. We want to emphasize that per experiment, both the multi RDM and the monolithic RDM have the same amount of trainable parameters. Since the monolithic RDM does not have a TCN and to maintain a fair comparison, the architecture of the monolithic RDM is slightly modified. Another network head is added and trained to output the current task ID similarly to the TCN in the multi RDM architecture. This way, the monolithic architecture gets exactly the same training information as the multi RDM architecture. Training is summarized as follows:

\begin{enumerate}
    \item Collect around 30\,000 (gridworld) / 200\,000 (vizdoom) transitions from each task by random action sampling. Since only complete trajectories are collected, the final number of collected transitions might end up slightly higher.
    \item Train the VQ-VAE on the collected samples for 200 (both) epochs.
    \item Train either the monolithic or the multi RDM predictors on the trajectories using the VQ-VAE to embed observations. Training is done for 150 (gridworld) / 500 (vizdoom) epochs.
    \item Perform planning utilizing the trained predictors. 
\end{enumerate}

When fully trained, we use the VQ-VAE and RDMs to perform imaginary rollouts as described in Section \ref{sec:planning}. Since we specifically want to assess the quality of the learned dynamics for both the monolithic and the multi RDM, we do not re-plan actions after each step. This way, groundtruth observation data would be injected into the system continuously, which potentially helps a suboptimal model. Instead, in the gridworld experiment our algorithm uses only the starting observation of the agent to perform a full planning routine until the end of the episode. This has the beneficial side effect of being considerably less time consuming than re-planning after every step. Note that the starting observations of the gridworld tasks are unique, so they should in theory provide sufficient information for the algorithm. In the vizdoom experiment, one of the two tasks features a dynamic environment that can change without  interaction of the agent. Thus, periodic re-planning is necessary to incorporate new information from the environment. In an effort to find a good balance between computational cost and planning accuracy, we use the first 30 actions from the planning procedure and then re-plans from there on.
 
To obtain the gridworld experiment data discussed in the next section, we repeated the above mentioned four-step process three times with different random seeds and averaged the results appropriately. To obtain the vizdoom experiment data, we did a single iteration of the above mentioned four-step process. We want to clarify that we trained one pair of dynamics models (a monolithic and a multi RDM) per experiment. This means, the dynamics models trained on the gridworld experiment did not receive any data from the vizdoom experiment and vice versa.

\section{Experiments \label{sec:experiments}}

\begin{figure}
\centering
\includegraphics[width=0.32\textwidth]{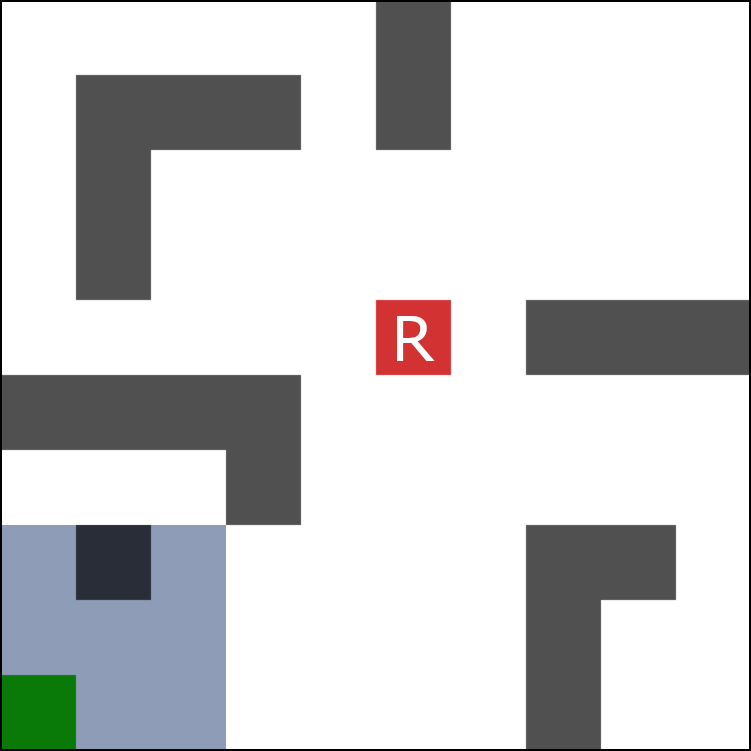}
\includegraphics[width=0.32\textwidth]{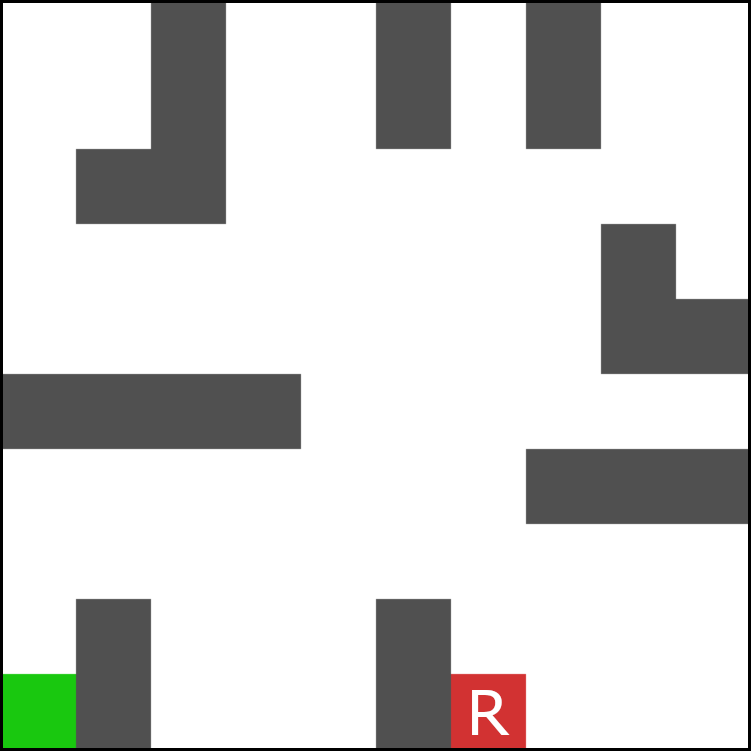}
\includegraphics[width=0.32\textwidth]{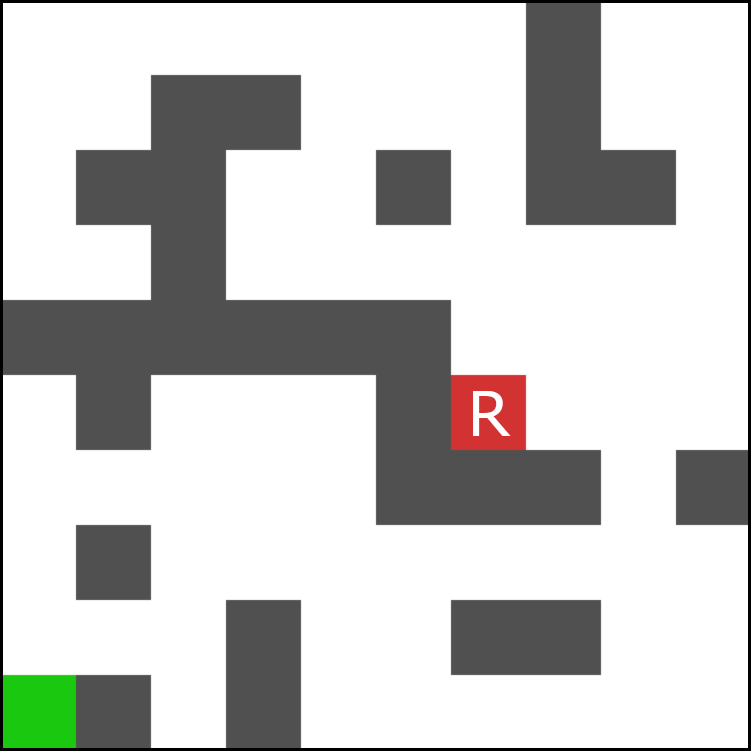}
\caption{The three training tasks of the gridworld experiment. The agent (green square in the lower left corner) has to find the reward (red square marked with "R"). Every action moves the agent one grid cell into the respective direction if no obstacles are met. Walls (black squares) block movement without reward penalty. The shaded blue area in the first environment (left panel) shows an example of the top-down viewport of the agent. If part of the viewport is outside of the environment at any time, the respective part is filled with zeros (i.e. it is colored black).}
\label{fig:environments_gridworld}
\end{figure}

\begin{figure}
\centering
\includegraphics[width=0.49\textwidth]{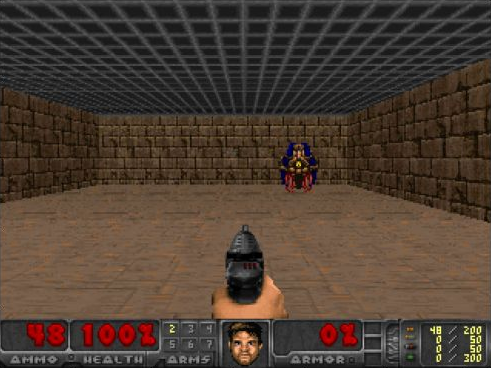}
\includegraphics[width=0.49\textwidth]{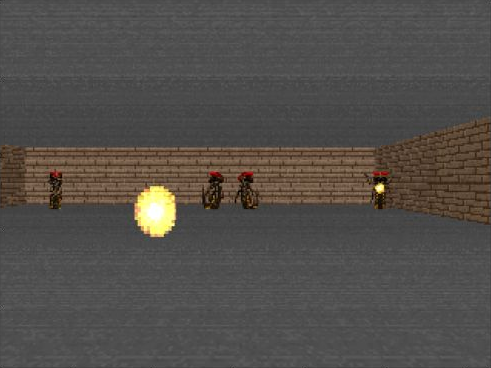}
\caption{The two training tasks of the vizdoom experiment. In both cases, the agent is situated in a rectangular room with its back against one wall, facing the opposite wall. The agent can move sideways left, sideways right and attack in both tasks. In the `VizdoomBasic'' task (left image), the agent has to shoot the monster on the opposite wall. In the ``TakeCover'' task (right image), the agent has to dodge incoming fireballs shot by enemies that randomly spawn on the opposite wall.}
\label{fig:environments_vizdoom}
\end{figure}

To answer the question of whether dynamics models can benefit from training on multiple tasks at once, we use a simple gridworld experiment (cf. Figure \ref{fig:environments_gridworld}) and a more complex vizdoom experiment (cf. Figure \ref{fig:environments_vizdoom}). For both experiments, observations from different tasks can easily be confused with each other. Yet, the system has to learn to identify the tasks and memorize task dynamics without confusing them. This is especially difficult for the gridworld experiment, since although having very different global structures, the individual image observations look similar across all three tasks. On the one hand, this makes it easier to learn shared properties across them, e.g. that all objects in the viewport are shifted down one cell if an ``up'' move is performed. We expect this to be advantageous for the monolithic RDM. On the other hand, in the absence of striking visual cues it becomes harder to distinguish between different tasks, which should put the multi RDM at an advantage because it uses the specialized TCN for that. In case of the vizdoom experiment, distinguishing the tasks is easier because there are salient optical cues in every observation (e.g. the gun and HUD indicating the ``VidzoomBasic'' task). We expect this to benefit the monolithic RDM. Yet, although the environment dynamics are partially similar for both tasks (agent movement), we see quite different observations and reward functions. We expect this reduced degree of common structure among the vizdoom tasks to put the multi RDM at an advantage.

\subsubsection{Gridworld Experiment}

All mazes have a size of 10 by 10 cells, a fixed layout per maze for reproducibility and a maximum episode length of 100 steps. Per environment, there exists one reward of value $1.0$ at a fixed position. Additionally, the agent receives a $-0.01$ penalty per step, resulting in a positive reward only if the agent is able to find the goal state fast enough. During data collection, the agent starts at a random, unobstructed cell somewhere in the maze and performs random actions. This takes exploration strategies, which exceed the scope of this work, out of the equation. To assess the performance of the learned dynamics models in a comparable way, the agent's starting position is fixed to the lower left corner of every maze during control. The action space is discrete and provides four choices: Up, down, left and right. The observation emitted by the environment is an image of a 5x5 cells top-down view centered on the agent position, visualized in Figure \ref{fig:environments_gridworld} in the leftmost panel. Although we conduct our experiments on a gridworld, we emphasize that the input space consists of proper image observations.

\subsubsection{Vizdoom Experiment}

In both tasks the agent is positioned in a rectangular 3D room viewed from the ego perspective (c.f. Figure \ref{fig:environments_vizdoom}). Standing with its back against one wall, the agent faces the opposite wall and can choose from three actions: Move sideways left, move sideways right and attack. Since the agent does not hold a weapon in the second task, the attack action here simply advances the environment one time step. In the first task (``VizdoomBasic''), the agent has to shoot the monster at the opposite wall for a reward of $100$. Every move action results in a reward penalty of $-1$ and every fired shot in a reward penalty of $-5$. The maximum number of allowed time steps is 300. In the second task (``TakeCover'') the agent has to evade fireballs shot by enemies randomly spawning at the opposite side of the room. Since every time step the agent is alive results in a reward of $1$, the goal is to survive as long as possible. Although there is formally no time limit for the second task, more and more monsters will spawn over time. This makes evading all fireballs increasingly difficult and ultimately impossible. For both tasks, the agent and monster starting positions are randomized during training data collection and fixed for the control experiments. Both, monolithic and multi RDM get the same sequence of starting positions.

\begin{figure}
\centering
\includegraphics[width=\textwidth]{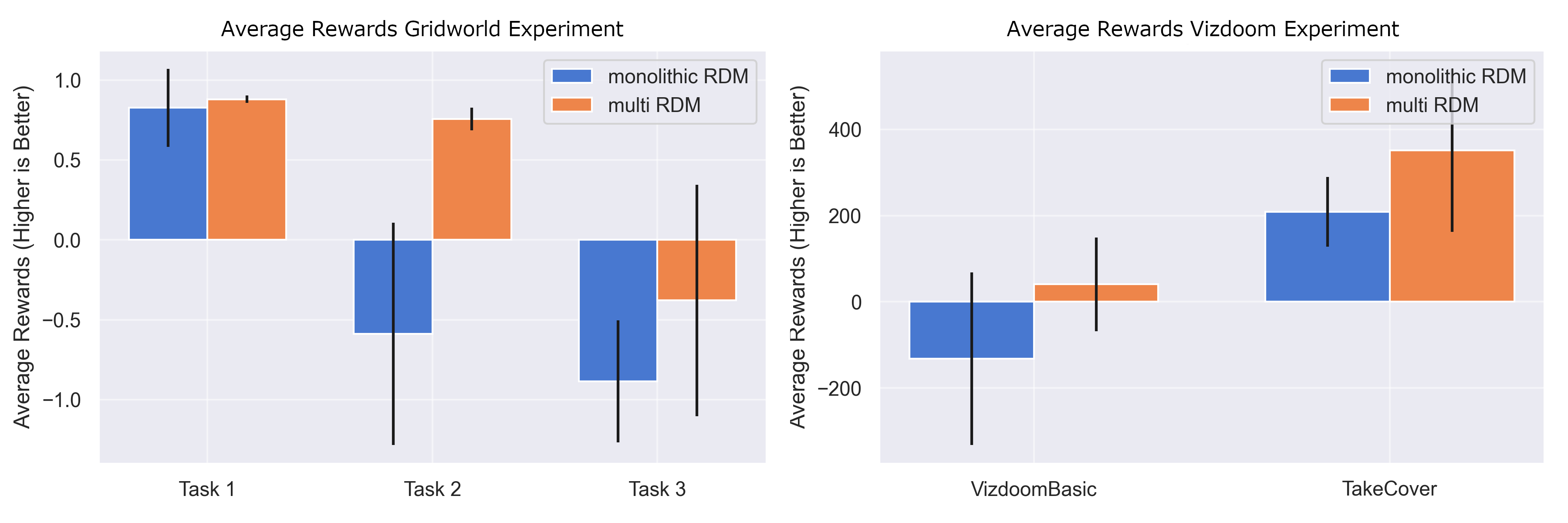}
\caption{Control results for the gridworld and the vizdoom experiments. Each task is depicted as a column on the x-axis, in turn separated into two columns representing the monolithic (blue) or multi (orange) RDM. The y-axis shows the average episode rewards. \textbf{Left:} Results for the gridworld experiment obtained via planning rollouts averaged over 60 trials per task. Note that the 60 trials originate from 3 full training runs as explained in Section \ref{sec:training} with 20 trials per run. \textbf{Right:} Results for the vizdoom experiment obtained via planning rollouts averaged over 35 trials per task. Note that the 35 trials originate from 1 full training run as explained in Section \ref{sec:training}.}
\label{fig:planning_results_both}
\end{figure}

\subsection{Evaluation}

To evaluate the quality of the trained monolithic as well as multi RDM, we use them to directly generate agent behavior via planning. This way, the performance of the approaches can be assessed with minimal additional complexity. The results obtained with our trained RDMs for the gridworld experiment are shown in Figure \ref{fig:planning_results_both} (left panel). The overall difficulty increase from task one to three is reflected in the reward decrease for both approaches. While the first task can be completed with a comparable performance by both architectures, the second and third task show significant differences between the two approaches, the split architecture outperforming the monolithic network ($p<0.0001$, Mann-Whitney-U-test). The third task is generally the hardest, since the way to the reward is most obstructed and longer than for the other tasks. 
The results for the vizdoom experiment are shown in Figure \ref{fig:planning_results_both} (right panel) and follow a similar pattern as those of the gridworld experiment. The multi RDM architecture manages to deliver significantly higher average rewards than the monolithic architecture in both tasks ($p<0.002$, Mann-Whitney-U-test). Note that the maximum reward that can be obtained in the first task is capped at around $85$ (depending on the agent and monster starting positions) while it is potentially infinite for the second task.

\begin{figure}
\centering
\includegraphics[width=0.32\textwidth]{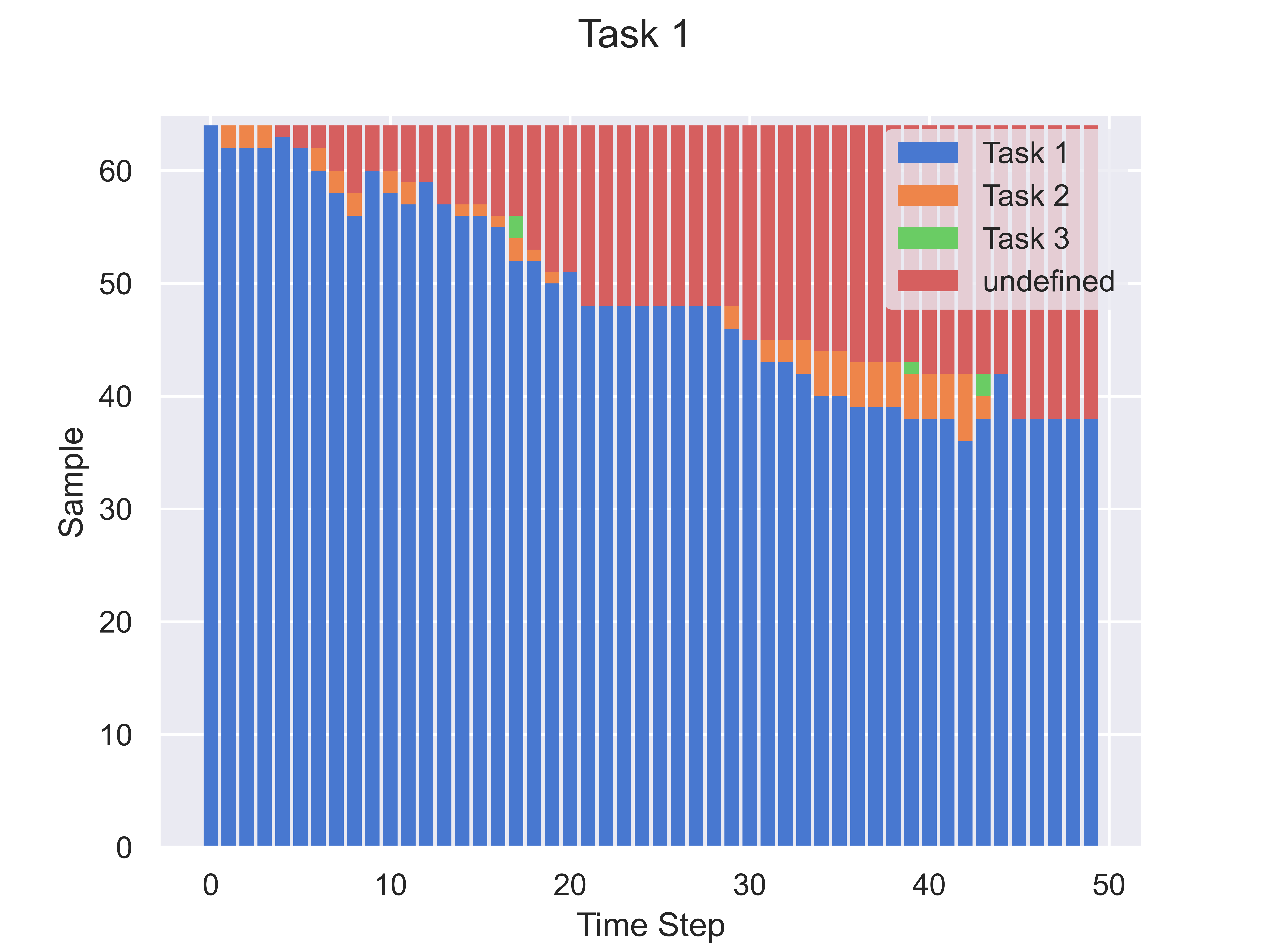}
\includegraphics[width=0.32\textwidth]{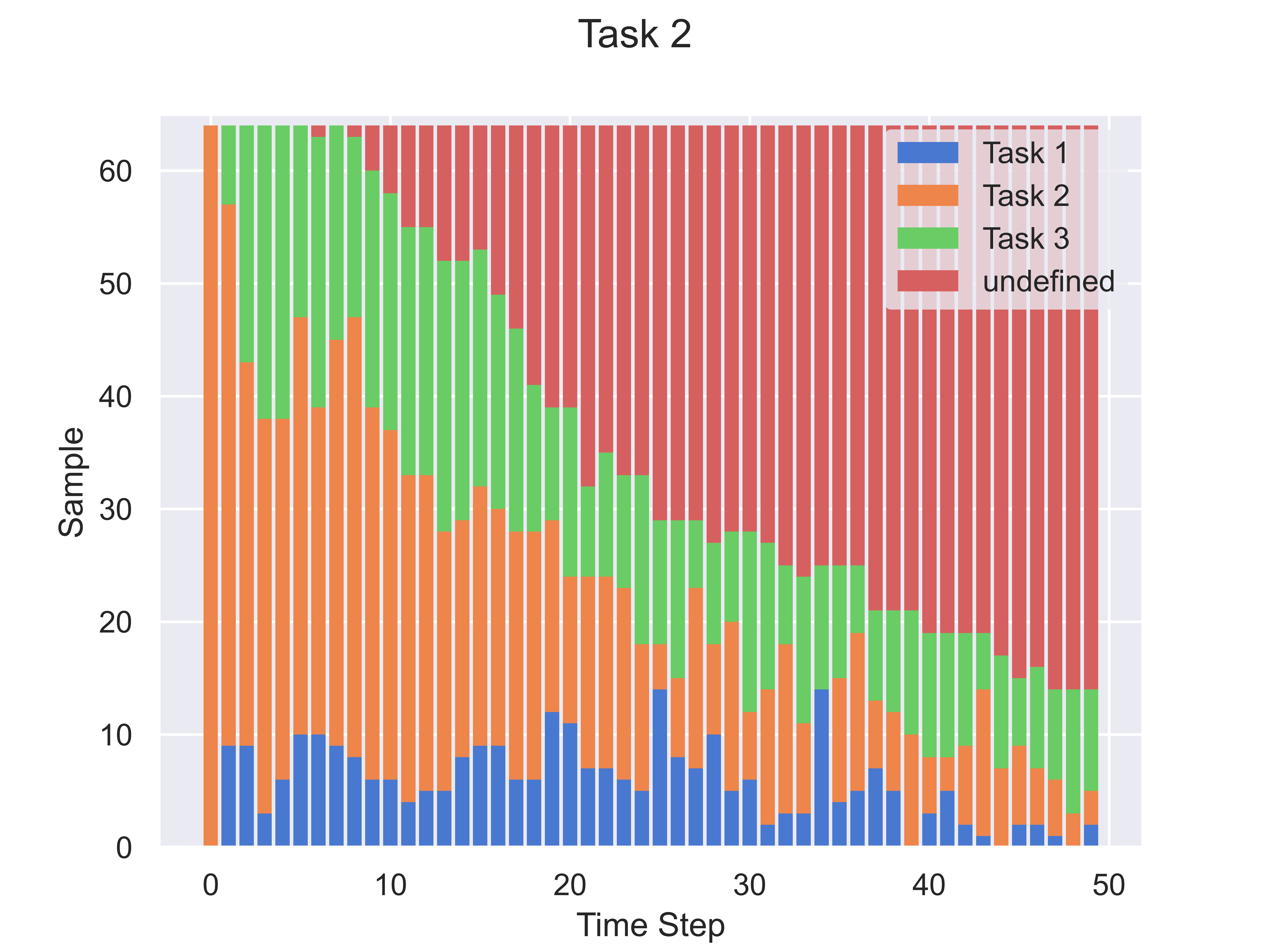}
\includegraphics[width=0.32\textwidth]{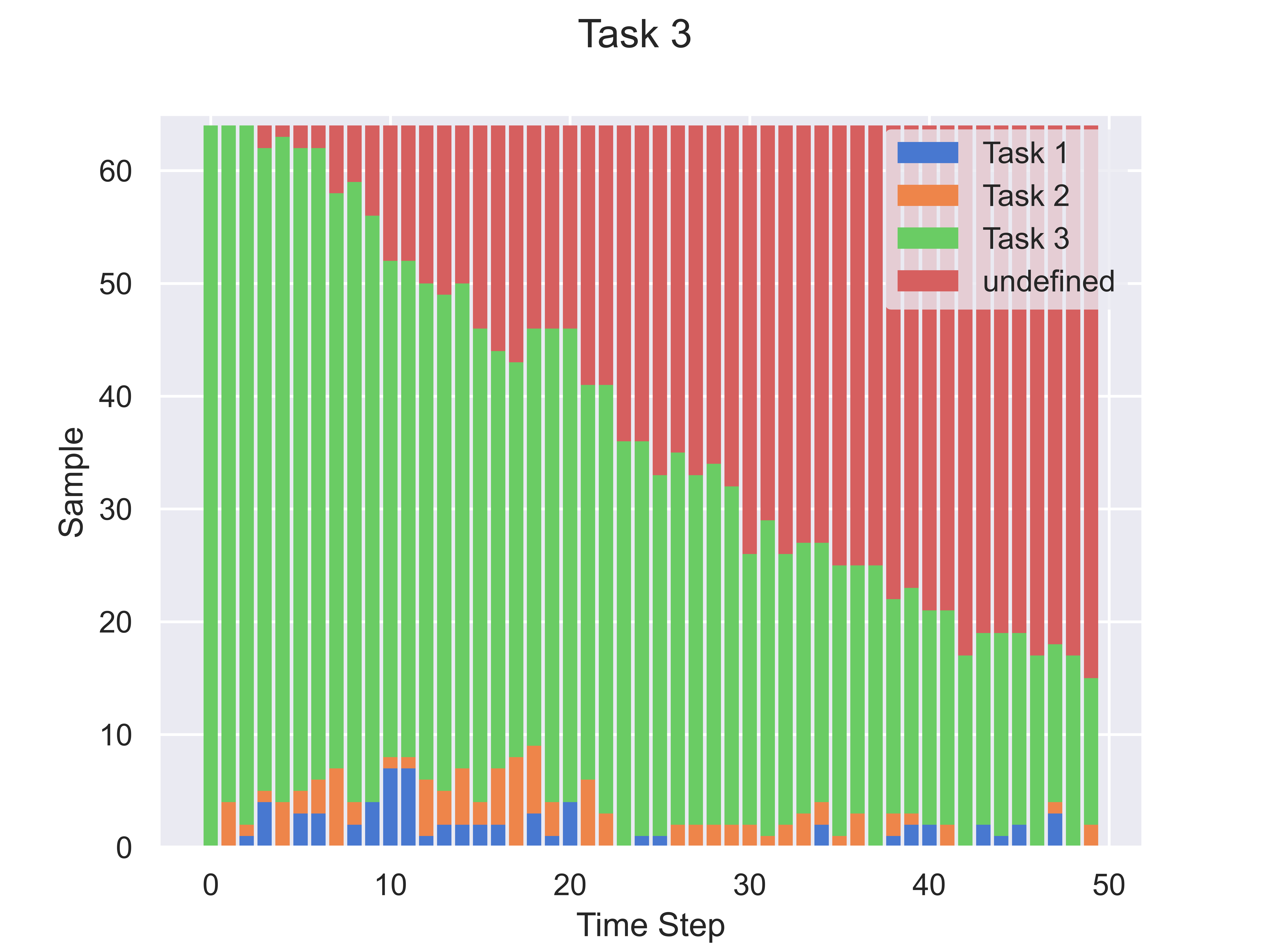}
\includegraphics[width=0.32\textwidth]{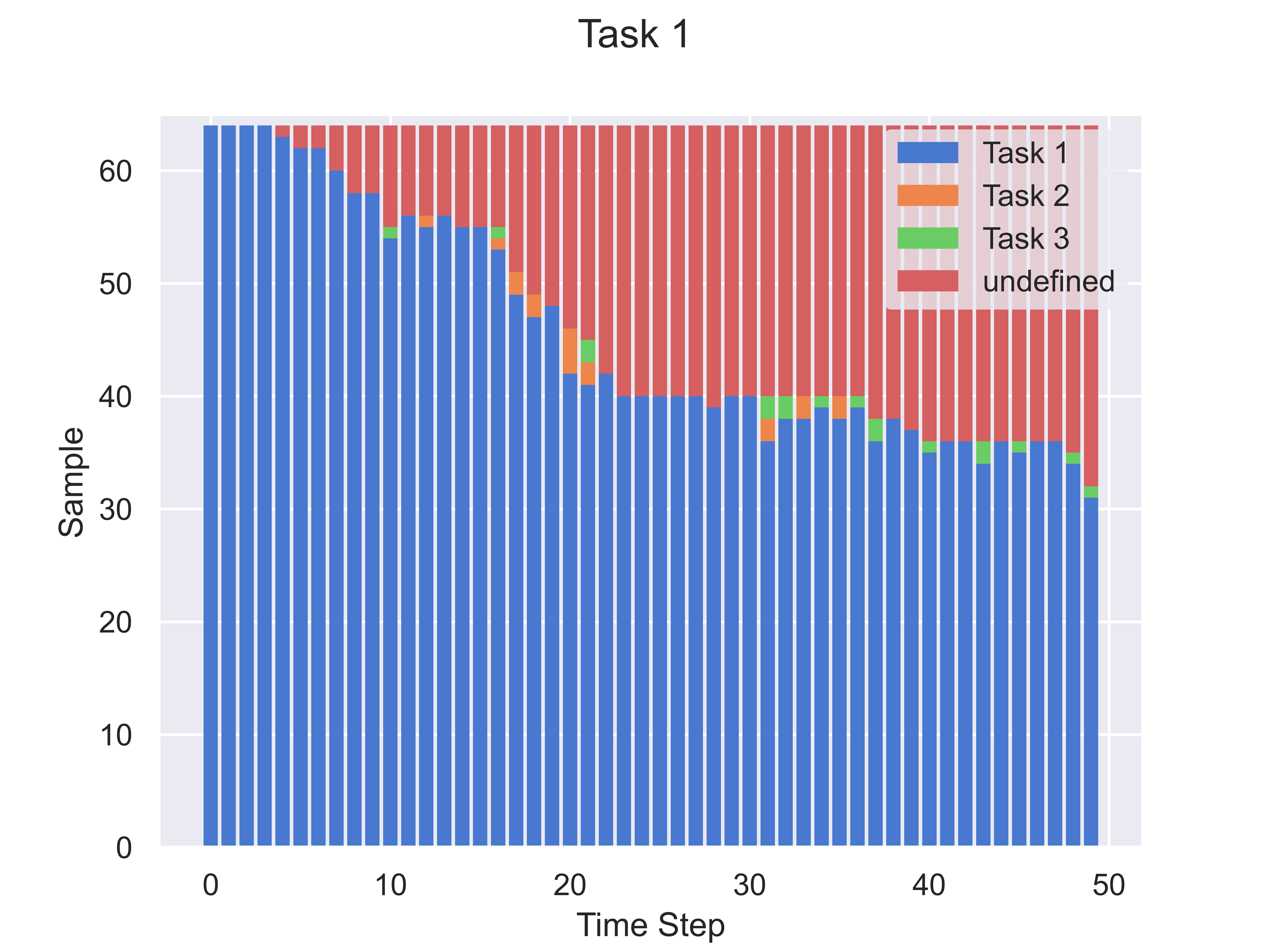}
\includegraphics[width=0.32\textwidth]{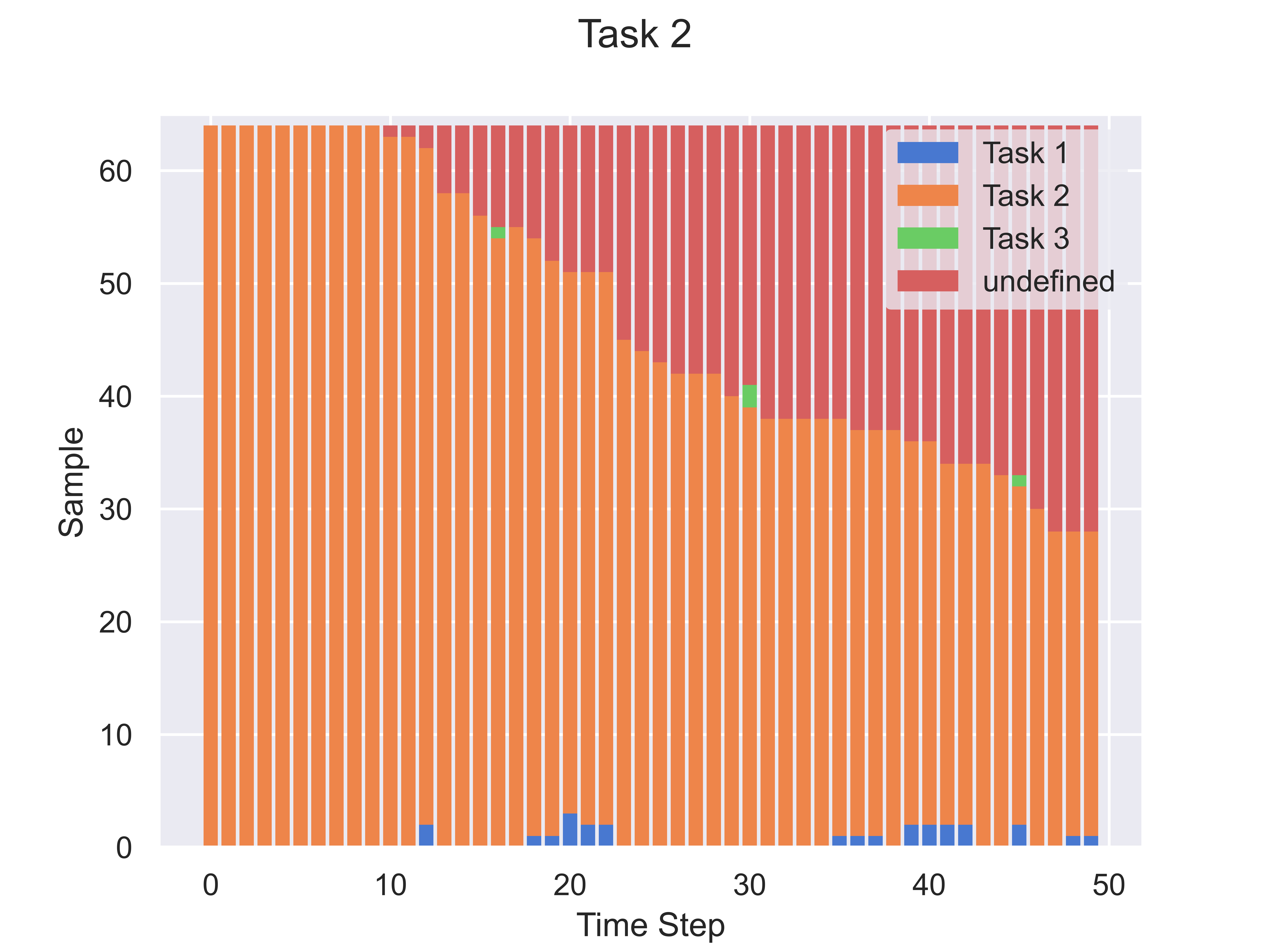}
\includegraphics[width=0.32\textwidth]{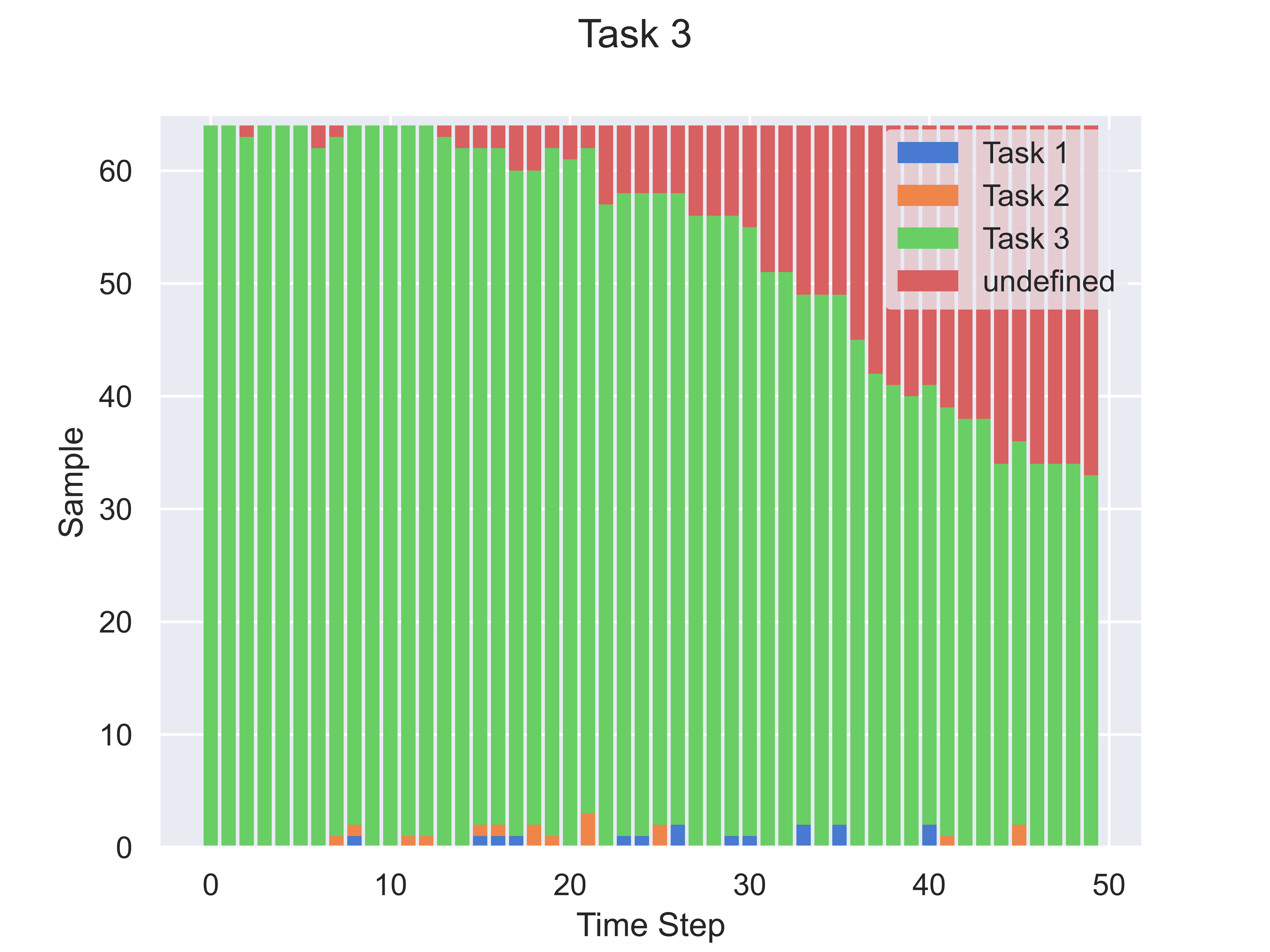}
\caption{For each plot, 64 rollouts of length 50 with randomly chosen initial observations and random actions were performed. On the y-axis all samples of a single time step are listed, color-coded depending on their most probable class. Samples belonging to task (1, 2, 3) are colored in (blue, orange, green) and samples that could not be assigned to any task are colored in red. On the x-axis, the distinct time steps are shown. For visual clarity, data obtained exclusively from one of the three training runs is shown. \textbf{Top row:} Results of the monolithic architecture. \textbf{Bottom row:} Results of the multi RDM.}
\label{fig:env_per_trajectory}
\end{figure}

With these findings in mind, the question is why the monolithic RDM performs worse than the multi RDM in planning. To shed light on possible reasons for the performance difference, we carefully inspected the generated rollout samples of both approaches in the gridworld experiment setting. Per task and architecture, we generated 64 trajectories of length 50 from 64 arbitrary starting observations. The actions were randomly sampled from the tasks' action spaces. Figure \ref{fig:env_per_trajectory} shows the most likely environment of the produced latent space embedding $z_{i, t}$ for every trajectory $i$ and time step $t$. To find the most likely environment, each $z_{i, t}$ was decoded into an image using the VQ-VAE decoder and compared to a list of representative images. This list contains one image for every possible agent location in the three tasks, which means that if the decoded $z_{i, t}$ resembles any valid observation, this was detected with absolute certainty. If the average per-pixel distance to the closest match was above a certain threshold (here $0.01$), the $z_{i, t}$ was classified as ``undefined'' (red in the plots). Otherwise, it was classified to belong to the environment of the closest match (blue for task 1, orange for task 2, green for task 3). Note that this analysis is feasible only for environments with a relatively small amount of states, which is another reason why we decided to use the gridworld tasks for this analysis.

It is clearly visible that the monolithic RDM produces trajectories of lower overall quality (cf. Figure \ref{fig:env_per_trajectory}, top row). First of all, compared to the multi RDM the fraction of undefined observations is notably higher in tasks 2 and 3. Even more importantly, the monolithic RDM has a higher chance to generate $z$ that belong to a different task, most evidently reflected in task 2 (cf. Figure \ref{fig:env_per_trajectory}, top row, middle panel). This can lead to malformed rollout trajectories which miss rewards or report rewards where there are none, harming the planning process in general. In the worst case, the monolithic RDM switches tasks in the middle of a rollout trajectory. By visual inspection of the rollout trajectories, we observed this to happen regularly for the monolithic RDM in both the gridworld and the vizdoom experiments. Through separation of the task dynamics in the multi RDM, considerably less confusion among the tasks arises and the fraction of undefined samples is reduced as well. This stands in line with measurably better planning performance for the multi RDM architecture. We conclude from our analysis that catastrophic interference outweighs possible advantages of transfer learning in the monolithic RDM. In our experiments, the simple use of isolated dynamics models mitigates the harmful interference to a large degree.

\section{Conclusion}

In this work, we demonstrate an unintuitive effect in model-based multi-task RL. Contrary to expectations fueled by model-free multi-task RL, using a single monolithic RDM for all tasks can harm performance in model-based multi-task RL instead of improving it. We conclusively assume that the positive effect of transfer learning between similar tasks is either absent for dynamics models or outweighed by catastrophic interference. Moreover, while using the same parameter budget for both approaches we show that imposing an internal structure to the dynamics model can lead to notably improved learned dynamics. We show this by using both approaches for control through planning, where the multi RDM scores measurably higher rewards than the monolithic RDM. By analyzing the rollout trajectories of both the monolithic and the multi RDM in the gridworld setting, we find cleaner, more task-specific trajectories and less overall wrong predictions in case of the multi RDM. We conclude that separating task recognition from task dynamics effectively prevents catastrophic interference (or task confusion) to a large degree, leading to the measured performance improvement. The code for reproducing our experiments, including all hyperparameters, can be found at \textit{https://github.com/rschiewer/lrdm}.
\\
\\
\noindent
\textbf{Acknowledgements:} We thank Jan Bollenbacher, Dr. Anand Subramoney and Prof. Dr. Tobias Glasmachers for their feedback and help, which greatly influenced this work. 
%
%
%
\bibliographystyle{splncs04}
\bibliography{main}
\end{document}